\pdfoutput=1
\documentclass[11pt]{article}

\usepackage[]{EMNLP2023}

\usepackage{times}
\usepackage{latexsym}
\usepackage{graphicx}
\usepackage{booktabs}
\usepackage{tabularx}
\usepackage{url}
\usepackage[T1]{fontenc}
\usepackage[utf8]{inputenc}
\usepackage{microtype}
\usepackage{inconsolata}

\usepackage{pdfpages}

\title{\href{https://sign.mt}{sign.mt}: Real-Time Multilingual Sign Language Translation Application}

\author{Amit Moryossef \\
  \texttt{amit@sign.mt} \\ \\
  If you want to go fast, go alone; \\
  If you want to go far, go together. }

\begin{document}
\maketitle

\begin{abstract}
This paper presents \textit{sign.mt}, an open-source application for real-time multilingual bi-directional translation between spoken and signed languages. 
Harnessing state-of-the-art open-source models, this tool aims to address the communication divide between the hearing and the deaf, facilitating seamless translation in both spoken-to-signed and signed-to-spoken translation directions. 

Promising reliable and unrestricted communication, \textit{sign.mt} offers offline functionality, crucial in areas with limited internet connectivity. It further enhances user engagement by offering customizable photo-realistic sign language avatars, thereby encouraging a more personalized and authentic user experience. 

Licensed under \href{https://creativecommons.org/licenses/by-nc-sa/4.0/}{CC BY-NC-SA 4.0}, \textit{sign.mt} signifies an important stride towards open, inclusive communication.
The app can be used, and modified for personal and academic uses, and even supports a translation API, fostering integration into a wider range of applications. However, it is by no means a finished product. 

We invite the NLP community to contribute towards the evolution of \textit{sign.mt}. Whether it be the integration of more refined models, the development of innovative pipelines, or user experience improvements, your contributions can propel this project to new heights. Available at \url{https://sign.mt}, it stands as a testament to what we can achieve together, as we strive to make communication accessible to all.
\end{abstract}

\section{Motivation}

Sign language translation applications are crucial tools for enabling communication between individuals who are deaf or hard of hearing and those who communicate through spoken language. However, the complexity of developing sign language translation applications goes beyond handling mere text. These applications must be able to process and generate videos, demanding additional considerations like compute capabilities, accessibility, usability, working with large files, and platform support.

\textit{sign.mt}, standing for \textbf{Sign} Language \textbf{M}achine \textbf{T}ranslation, was conceived as a response to these challenges. Current research in the field of sign language translation is fragmented and somewhat nebulous, with different research groups focusing on various aspects of the translation pipeline or on specific languages. Moreover, the high costs associated with server-side deployment and the complexity of client-side implementations often deter the development of interactive demonstrations for newly proposed models.

By providing a comprehensive application infrastructure that integrates the essential features around the translation process, \textit{sign.mt} serves as a dynamic proof-of-concept. 
It aims to streamline the integration of new research findings into the application, sidestepping the overhead typically associated with implementing a full-stack application.
When a research group develops a new model or improves a pipeline, they can integrate their advancements into the app swiftly, focusing only on their model. This approach allows researchers to deploy the app in a branch, testing their models in a practical environment. If the license allows and the models show an improvement, they can contribute their models to the main codebase.
This is the first tool of its kind, diverging significantly from closed-source commercial applications.

Further, \textit{sign.mt} serves as a multilingual platform, thus unifying the fragmented research landscape. It enables the concurrent running of models from different research groups for the supported languages, providing users with state-of-the-art translation capabilities for each language. Through this, \textit{sign.mt} not only enhances accessibility and communication but also fuels continuous innovation in sign language translation research.

\section{Implementation}

Sign language translation presents unique challenges that set it apart from text-based translation. 
While text-based translation operates entirely within the textual domain for both input and output, sign language translation involves cross-modal transformation – from text to video and vice versa. This demands distinct implementations not only in functionality but also in the user interface.

It is essential to emphasize that the specific models utilized within various pipelines are deliberately modular and interchangeable. Our current choice of models for each module or task is primarily opportunistic, driven by availability rather than performance metrics or user evaluations. The app serves as a dynamic orchestrator, seamlessly coordinating among these models to deliver an integrated user experience. The platform's design accommodates the likelihood that researchers or users may wish to experiment with different models or fine-tune existing pipelines, without being constrained by rigid implementation details.

\subsection{Spoken-to-Signed Translation}

\begin{figure}[t]
\centering
\includegraphics[width=\linewidth]{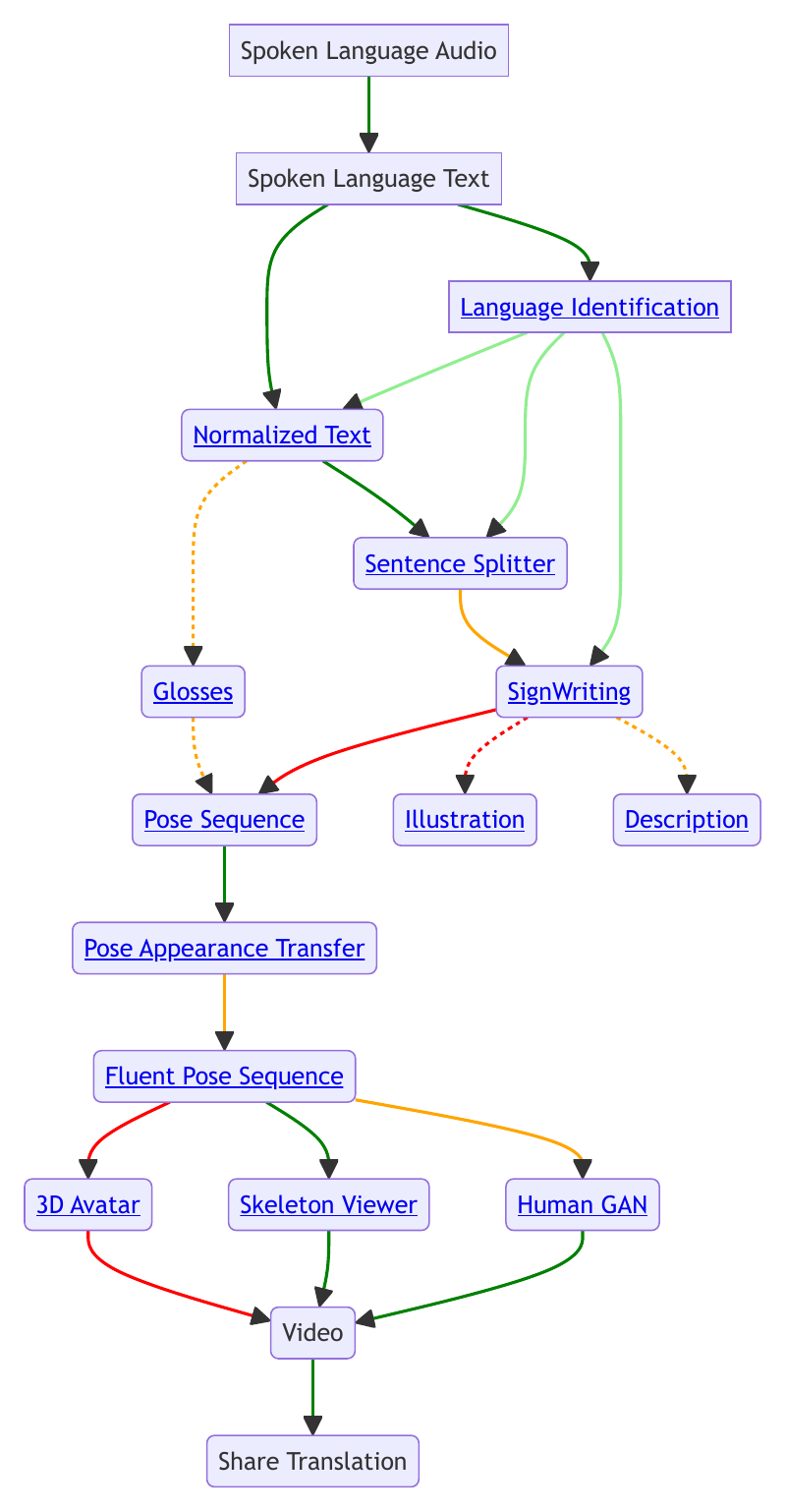}
\caption{The Spoken-to-Signed translation pipeline.}
\label{fig:implementation-spoken-to-signed}
\end{figure}

Through this pipeline (Figure \ref{fig:implementation-spoken-to-signed}), \textit{sign.mt} is capable of real-time translation from spoken language audio (or text) into sign language video, further democratizing communication across modalities.

% Audio-to-Text
For spoken-to-signed translation, the process begins with an input of spoken language text. Optionally, we allow audio input, which is first transcribed into spoken language text using on-device Speech-to-Text (STT) technology.

% Language Identification, Normalization, and splitting
When the input language is unknown, the text undergoes Spoken Language Identification (using \texttt{MediaPipe} \cite{mediapipe} or \texttt{cld3} \cite{cld3}), which detects the language of the provided text. This is crucial for choosing the appropriate model for subsequent translation steps. Simultaneously, the text is optionally normalized (using \texttt{ChatGPT} \cite{openai2023chatgpt}). This includes fixing capitalization, punctuation, grammatical errors, or misspellings, which we have found to enhance the performance of subsequent translation stages.
The language-identified and potentially normalized text is then split into individual sentences using the on-device internationalized segmentation service \cite{mdn_intl_segmenter}. 

% Text to SignWriting translation
Each sentence is then individually translated into SignWriting \cite{writing:sutton1990lessons}. Here, our system leverages real-time client-side machine translation \cite{bogoychev-etal-2021-translatelocally} to translate the grammatical structures and lexicon of spoken languages into the visual-gestural modality of sign languages \cite{jiang2022machine,moryossef2023signbank}.

% Animation, Pose anonymization, Rendering
The SignWriting output is then converted into a pose sequence (Inspired by \citet{shalev2022ham2pose}), representing the signed sentence. After undergoing appearance transfer to always show the same person \cite{moryossef2024anonymization}, this pose sequence is the input for the rendering engine, with three options: Skeleton Viewer (Minimalistic visualization of the skeletal pose \cite{moryossef2021pose-format}) Human GAN (Pix2Pix \cite{isola2017image,Shi2016RealTimeSI} image-to-image model, generating a realistic human avatar video), and a 3D Avatar (Neural model to translate between pose positions and rigged rotations, performing the signs).

These different outputs provide users with a choice on how they prefer to view the translation, catering to a broad range of preferences and use cases. The skeleton viewer is useful for developers to see the raw output, as well as for low-compute users. The 3D Avatar is useful in mixed reality applications, where it can be integrated in the environment, and the Human GAN is useful for high-compute users, facilitating a natural interaction.

Currently, while we don't have a fully functional SignWriting to pose animation model, we have created a baseline model as an interim solution \cite{moryossef2023baseline}. This model performs dictionary-based translation from the spoken language text directly to poses, bypassing the SignWriting stage. However, it's important to note that there are numerous common cases in sign languages that this baseline model cannot handle adequately yet. We have made the baseline model open-source, and it is available for further improvements and contributions from the community at \url{https://github.com/sign-language-processing/spoken-to-signed-translation}. We hope that this open-source approach will stimulate further research and development in this area, allowing for the integration of more sophisticated and accurate models in future iterations of the application.

\subsection{Signed-to-Spoken Translation}

\begin{figure}[t]
\centering
\includegraphics[width=\linewidth]{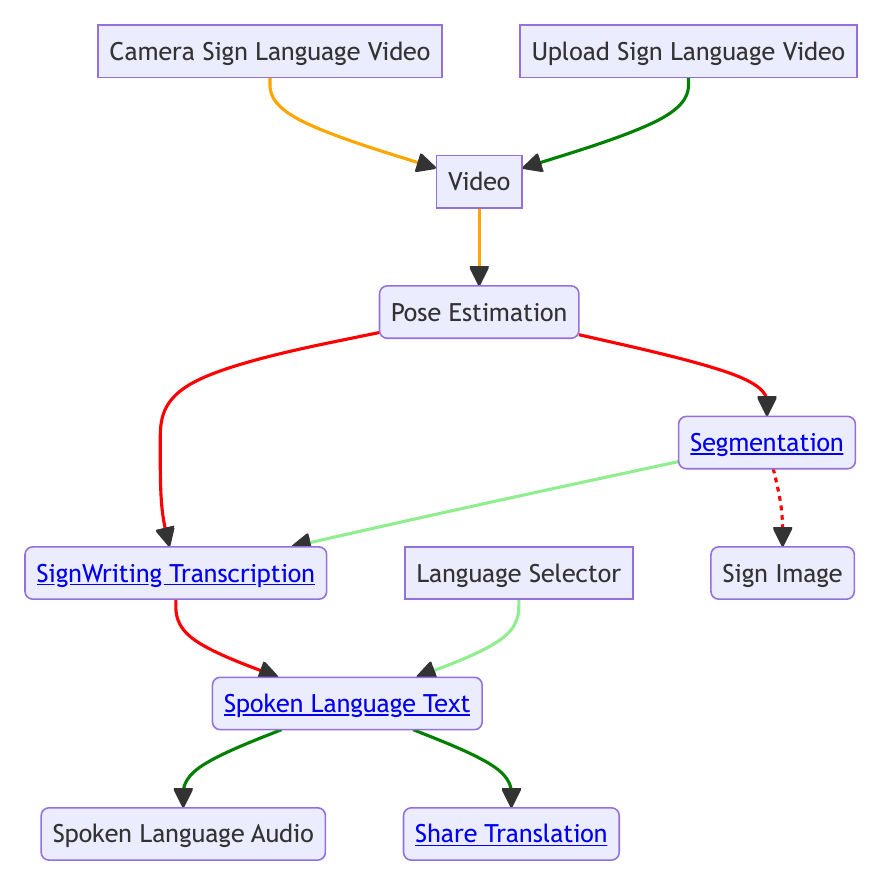}
\caption{The Signed-to-Spoken translation pipeline.}
\label{fig:implementation-signed-to-spoken}
\end{figure}

Through this pipeline (Figure \ref{fig:implementation-signed-to-spoken}), \textit{sign.mt} can take a sign language video and output corresponding spoken language text or audio in real-time. The offline functionality of the app ensures that this feature remains accessible even in areas with limited connectivity, provided that the models are cached on the device.

For signed-to-spoken translation, the source is a video (either by the user uploading a pre-existing sign language video or using the camera to record a live sign language video). 
Our current pipeline takes the video, and using Mediapipe Holistic \citep{mediapipe2020holistic} pose estimation extracts the full body pose from each frame.

This pose information is then fed into a Segmentation module \cite{moryossef2023segmentation}, which segments distinct signs within the continuous signing flow, as well as phrase boundaries. The segmented signs are subsequently lexically transcribed using SignWriting \citep{writing:sutton1990lessons}, a comprehensive system for transcribing sign languages visually.

This SignWriting transcription serves as the textual input for the translation model, which translates it into corresponding spoken language text \citep{jiang2022machine,moryossef2023signbank}. This text is then optionally converted into spoken language audio using on-device Text-to-Speech (TTS), providing an auditory output for the user.

\section{User Engagement}

The impact of \textit{sign.mt} can be measured by its widespread and consistent usage, highlighting the tremendous growth potential as the app continues to slowly improve. 

\begin{figure}[h]
\centering
\includegraphics[width=\linewidth]{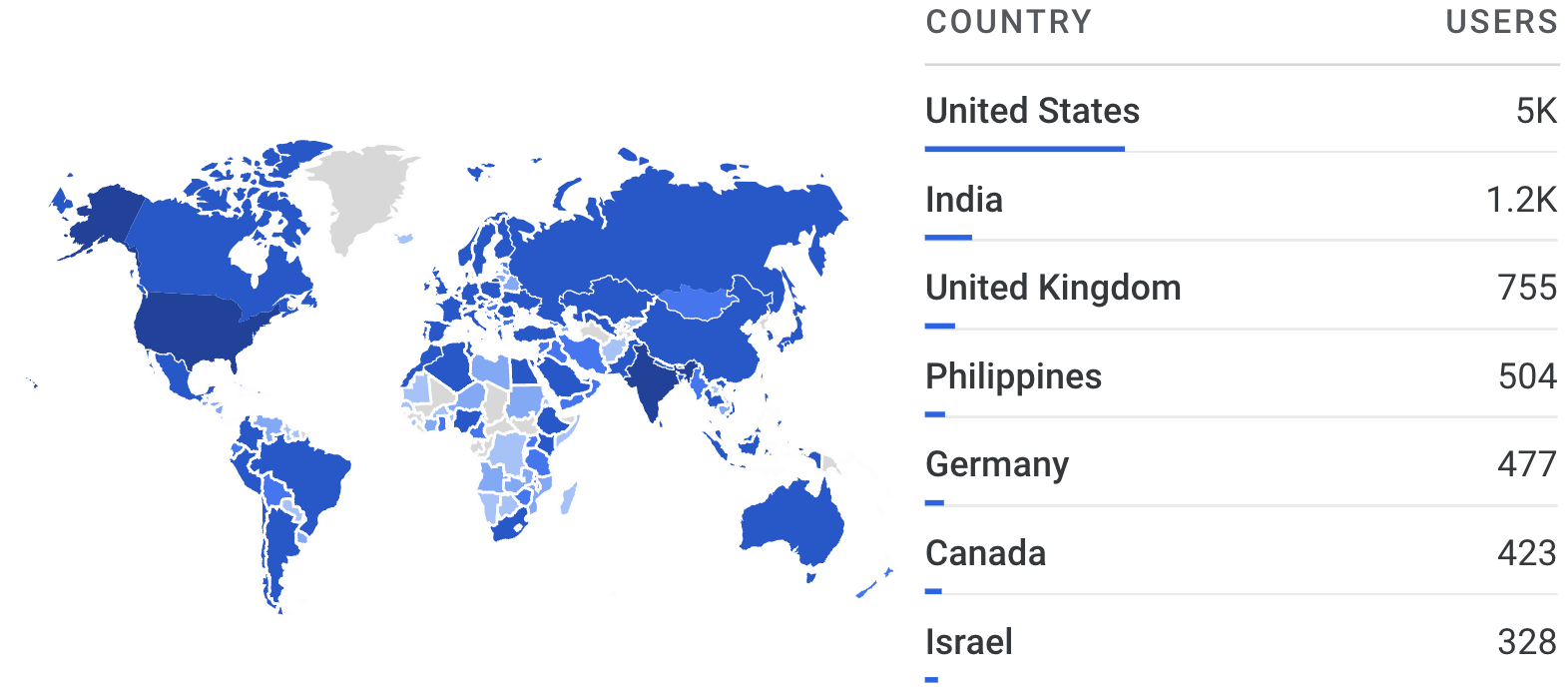}
\caption{Distribution of \textit{sign.mt} users across the world, over the last year.}
\label{fig:users-by-country}
\end{figure}

Figure~\ref{fig:users-by-country} depicts the global adoption of \textit{sign.mt}, with users distributed across multiple countries. None of these top user countries are home to the core developer of the app.

\begin{figure}[h]
\centering
\includegraphics[width=\linewidth]{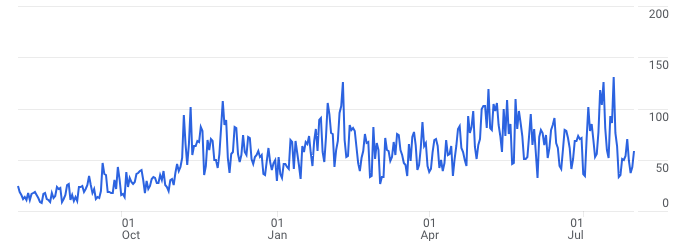}
\caption{Growth of \textit{sign.mt} users over the last year.}
\label{fig:users-over-time}
\end{figure}

As shown in Figure~\ref{fig:users-over-time}, \textit{sign.mt} demonstrates slow but consistent user growth (by Google Analytics), indicative of its reliability and sustained relevance.

\begin{figure}[h]
\centering
\includegraphics[width=\linewidth]{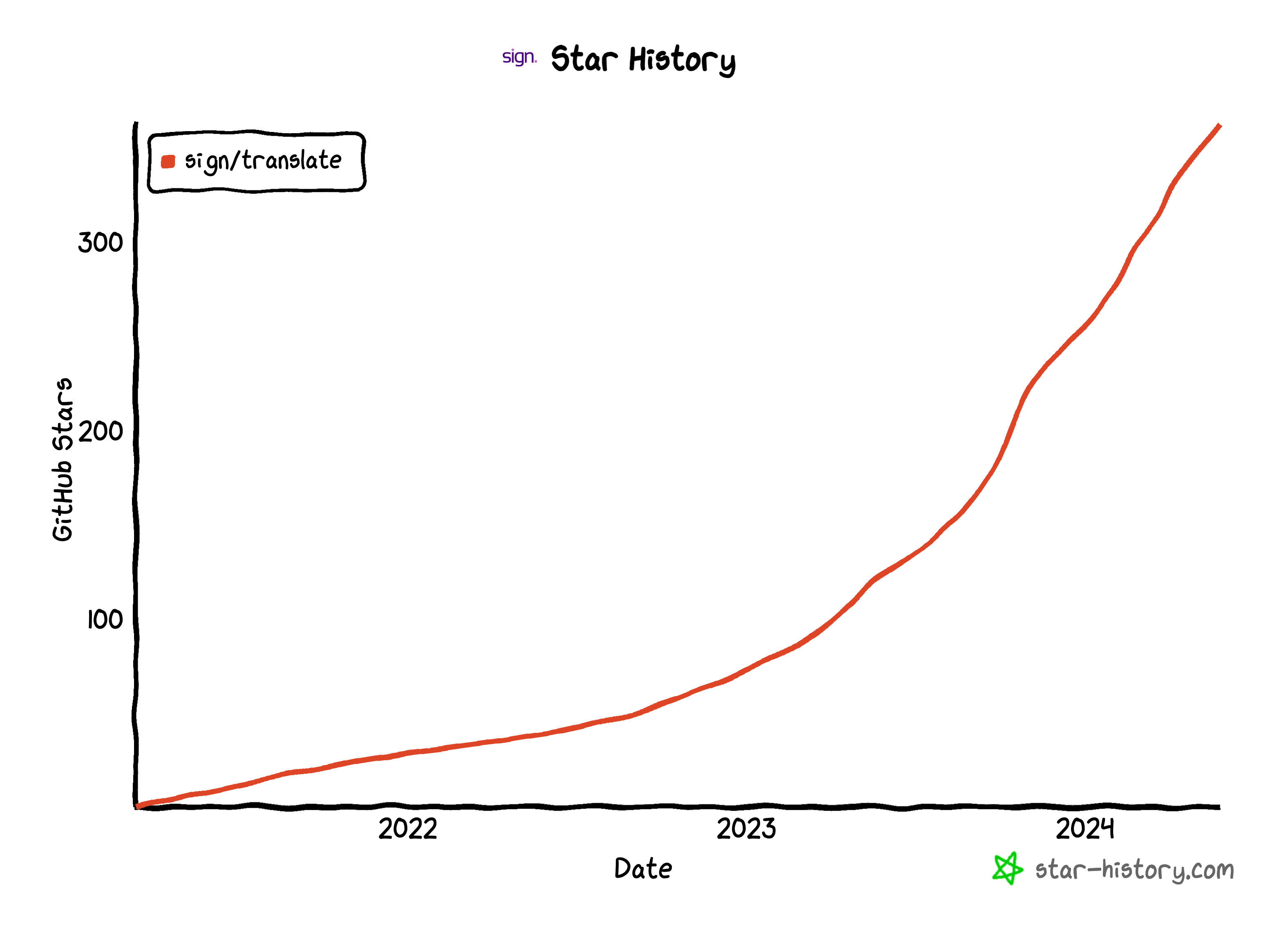}
\caption{Number of stars for the repository over time.}
\label{fig:stars-over-time}
\end{figure}

Further validation of the community interest in \textit{sign.mt} is evidenced by the increasing number of stars for its repository, reaching 363 stars as of May 25th, 2024 (Figure~\ref{fig:stars-over-time}).

\begin{figure}[h]
\centering
\includegraphics[width=\linewidth]{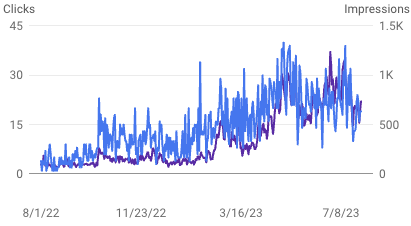}
\caption{Google Search Console metrics showing increasing interest in \textit{sign.mt}. (Clicks in blue)}
\label{fig:search-over-time}
\end{figure}

Public interest in \textit{sign.mt} is further supported by Google Search Console metrics (Figure~\ref{fig:search-over-time}), showing a significant increase in impressions and clicks over the past six months: 3.75K clicks (up from 1.56K), and 106K impressions (up from 24.4K). Despite the absence of a marketing team and a single maintainer, \textit{sign.mt} has managed to carve a niche for itself in the realm of NLP tools, reiterating its significance and impact.

\section{Distribution}

The code for \textit{sign.mt} is openly accessible and available for contribution on GitHub at \url{https://github.com/sign/translate}, under CC BY-NC-SA 4.0. Open sourcing with a permissive license encourages the continuous refinement and enhancement of the app through contributions from the wider developer and research communities.

The web application is freely accessible at \url{https://sign.mt}, designed with a responsive layout to cater to both desktop and mobile devices. Adhering to the design principles native to each platform, the application ensures an intuitive and user-friendly experience across all devices.
With localization being a critical aspect of accessibility, the app interface supports 104 languages. Contributors can add their language or enhance the support for existing languages.

In addition to the web application, native builds for iOS and Android devices are also provided through the GitHub repository. While these are currently in development, the plan is to make them available on the respective app stores as they reach stability, thereby extending the reach of \textit{sign.mt} to a wider audience.

\section*{Limitations}

As an evolving open-source project, \textit{sign.mt} still faces several challenges and limitations. 

At present, the app does not provide complete support for every component of the translation pipeline. Notably, the SignWriting-to-pose animation model does not currently exist, and instead, we use a simple dictionary lookup approach \cite{moryossef2023baseline}. Although it serves as an interim solution, it is insufficient for handling signed languages. We eagerly anticipate and encourage contributions from the research community to fill this gap with more advanced models. 

Although the app aspires to be a multilingual platform, the availability of models for different languages is currently fragmented. We rely on the research community to develop and contribute models for different languages. The support for each language, therefore, depends on the respective models available, leading to varying degrees of effectiveness across languages. For example, the SignWriting translation module works reasonably well for English/American Sign Language, German/German Sign Language and Portuguese/Brazilian Sign Language translations, and much worse for all other language pairs. Another example is the dictionary-based baseline only working on languages where dictionaries are available.

Due to the client-side deployment, we are restricted to using relatively smaller models. This inevitably leads to trade-offs in terms of translation accuracy and quality. While the offline functionality ensures accessibility in low connectivity areas, the constraint on model size is challenging.

The video processing components, including pose estimation and video rendering, are computationally intensive. This demands significant computational power, limiting the app's performance on devices with lesser computing capabilities. Optimizing these components further to ensure a smoother user experience across a wider range of devices is a challenge, often met with using lower-end models to achieve smoothness at the cost of accuracy.

Despite these limitations, \textit{sign.mt} serves as a robust foundation upon which future advancements can be built. It continues to evolve in response to the feedback of the wider community, consistently striving towards the goal of facilitating accessible, inclusive communication.

\bibliography{anthology,custom}

\begin{thebibliography}{16}
\expandafter\ifx\csname natexlab\endcsname\relax\def\natexlab#1{#1}\fi

\bibitem[{Arkushin et~al.(2023)Arkushin, Moryossef, and
  Fried}]{shalev2022ham2pose}
Rotem~Shalev Arkushin, Amit Moryossef, and Ohad Fried. 2023.
\newblock {Ham2Pose}: Animating sign language notation into pose sequences.
\newblock pages 21046--21056.

\bibitem[{Bogoychev et~al.(2021)Bogoychev, Van~der Linde, and
  Heafield}]{bogoychev-etal-2021-translatelocally}
Nikolay Bogoychev, Jelmer Van~der Linde, and Kenneth Heafield. 2021.
\newblock \href {https://doi.org/10.18653/v1/2021.emnlp-demo.20}
  {{T}ranslate{L}ocally: Blazing-fast translation running on the local {CPU}}.
\newblock In \emph{Proceedings of the 2021 Conference on Empirical Methods in
  Natural Language Processing: System Demonstrations}, pages 168--174, Online
  and Punta Cana, Dominican Republic. Association for Computational
  Linguistics.

\bibitem[{Grishchenko and Bazarevsky(2020)}]{mediapipe2020holistic}
Ivan Grishchenko and Valentin Bazarevsky. 2020.
\newblock \href {https://google.github.io/mediapipe/solutions/holistic.html}
  {Mediapipe holistic}.

\bibitem[{Isola et~al.(2017)Isola, Zhu, Zhou, and Efros}]{isola2017image}
Phillip Isola, Jun-Yan Zhu, Tinghui Zhou, and Alexei~A Efros. 2017.
\newblock Image-to-image translation with conditional adversarial networks.
\newblock In \emph{Proceedings of the IEEE conference on computer vision and
  pattern recognition}, pages 1125--1134.

\bibitem[{Jiang et~al.(2023)Jiang, Moryossef, M{\"u}ller, and
  Ebling}]{jiang2022machine}
Zifan Jiang, Amit Moryossef, Mathias M{\"u}ller, and Sarah Ebling. 2023.
\newblock \href {https://aclanthology.org/2023.findings-eacl.127} {Machine
  translation between spoken languages and signed languages represented in
  {S}ign{W}riting}.
\newblock In \emph{Findings of the Association for Computational Linguistics:
  EACL 2023}, pages 1661--1679, Dubrovnik, Croatia. Association for
  Computational Linguistics.

\bibitem[{Lugaresi et~al.(2019)Lugaresi, Tang, Nash, McClanahan, Uboweja, Hays,
  Zhang, Chang, Yong, Lee, Chang, Hua, Georg, and Grundmann}]{mediapipe}
Camillo Lugaresi, Jiuqiang Tang, Hadon Nash, Chris McClanahan, Esha Uboweja,
  Michael Hays, Fan Zhang, Chuo-Ling Chang, Ming Yong, Juhyun Lee, Wan-Teh
  Chang, Wei Hua, Manfred Georg, and Matthias Grundmann. 2019.
\newblock \href
  {https://mixedreality.cs.cornell.edu/s/NewTitle_May1_MediaPipe_CVPR_CV4ARVR_Workshop_2019.pdf}
  {Mediapipe: A framework for perceiving and processing reality}.
\newblock In \emph{Third Workshop on Computer Vision for AR/VR at IEEE Computer
  Vision and Pattern Recognition (CVPR) 2019}.

\bibitem[{Moryossef(2024)}]{moryossef2024anonymization}
Amit Moryossef. 2024.
\newblock pose-anonymization: Remove identifying information from sign language
  poses.
\newblock \url{https://github.com/sign-language-processing/pose-anonymization}.

\bibitem[{Moryossef and Jiang(2023)}]{moryossef2023signbank}
Amit Moryossef and Zifan Jiang. 2023.
\newblock \href {http://arxiv.org/abs/2309.11566} {Signbank+: Multilingual sign
  language translation dataset}.

\bibitem[{Moryossef et~al.(2023{\natexlab{a}})Moryossef, Jiang, M{\"u}ller,
  Ebling, and Goldberg}]{moryossef2023segmentation}
Amit Moryossef, Zifan Jiang, Mathias M{\"u}ller, Sarah Ebling, and Yoav
  Goldberg. 2023{\natexlab{a}}.
\newblock \href {https://aclanthology.org/volumes/2023.findings-emnlp/}
  {Linguistically motivated sign language segmentation}.
\newblock In \emph{Findings of the Association for Computational Linguistics:
  EMNLP 2023}. Association for Computational Linguistics.

\bibitem[{Moryossef and M\"{u}ller(2021)}]{moryossef2021pose-format}
Amit Moryossef and Mathias M\"{u}ller. 2021.
\newblock pose-format: Library for viewing, augmenting, and handling .pose
  files.
\newblock \url{https://github.com/sign-language-processing/pose}.

\bibitem[{Moryossef et~al.(2023{\natexlab{b}})Moryossef, M{\"u}ller,
  G{\"o}hring, Jiang, Goldberg, and Ebling}]{moryossef2023baseline}
Amit Moryossef, Mathias M{\"u}ller, Anne G{\"o}hring, Zifan Jiang, Yoav
  Goldberg, and Sarah Ebling. 2023{\natexlab{b}}.
\newblock \href {https://github.com/ZurichNLP/spoken-to-signed-translation} {An
  open-source gloss-based baseline for spoken to signed language translation}.
\newblock In \emph{2nd International Workshop on Automatic Translation for
  Signed and Spoken Languages (AT4SSL)}.
\newblock Available at: \url{https://arxiv.org/abs/2305.17714}.

\bibitem[{{Mozilla Developer Network}(2020)}]{mdn_intl_segmenter}
{Mozilla Developer Network}. 2020.
\newblock \href
  {https://developer.mozilla.org/en-US/docs/Web/JavaScript/Reference/Global_Objects/Intl/Segmenter}
  {\emph{Intl.Segmenter}}.
\newblock
  \url{https://developer.mozilla.org/en-US/docs/Web/JavaScript/Reference/Global_Objects/Intl/Segmenter}.

\bibitem[{OpenAI(2022)}]{openai2023chatgpt}
OpenAI. 2022.
\newblock \href {https://openai.com/blog/chatgpt} {Chatgpt: Optimizing language
  models for dialogue}.

\bibitem[{Salcianu et~al.(2016)Salcianu, Golding, Bakalov, Alberti, Andor,
  Weiss, Pitler, Coppola, Riesa, Ganchev, Ringgaard, Hua, McDonald, Petrov,
  Istrate, and Koo}]{cld3}
Alex Salcianu, Andy Golding, Anton Bakalov, Chris Alberti, Daniel Andor, David
  Weiss, Emily Pitler, Greg Coppola, Jason Riesa, Kuzman Ganchev, Michael
  Ringgaard, Nan Hua, Ryan McDonald, Slav Petrov, Stefan Istrate, and Terry
  Koo. 2016.
\newblock Compact language detector v3 (cld3).
\newblock \url{https://github.com/google/cld3}.
\newblock Accessed: 2023-08-01.

\bibitem[{Shi et~al.(2016)Shi, Caballero, Husz{\'a}r, Totz, Aitken, Bishop,
  Rueckert, and Wang}]{Shi2016RealTimeSI}
Wenzhe Shi, Jose Caballero, Ferenc Husz{\'a}r, Johannes Totz, Andrew~P. Aitken,
  Rob Bishop, Daniel Rueckert, and Zehan Wang. 2016.
\newblock Real-time single image and video super-resolution using an efficient
  sub-pixel convolutional neural network.
\newblock \emph{2016 IEEE Conference on Computer Vision and Pattern Recognition
  (CVPR)}, pages 1874--1883.

\bibitem[{Sutton(1990)}]{writing:sutton1990lessons}
Valerie Sutton. 1990.
\newblock \emph{Lessons in sign writing}.
\newblock SignWriting.

\end{thebibliography}
\bibliographystyle{acl_natbib}

\end{document}